\documentclass[a4paper, 10pt, conference]{ieeeconf}      
\usepackage{FG2021}

\FGfinalcopy 
\usepackage{graphicx}

\IEEEoverridecommandlockouts                              
\usepackage{amsmath} 
\usepackage{amssymb}  
\usepackage{nicefrac}
\usepackage{xcolor}
\definecolor{SM_color}{rgb}{0.016, 0.600, 0.831}

\def\FGPaperID{7} 

\title{\LARGE \bf
Identity-Expression Ambiguity in 3D Morphable Face Models
}

\author{\parbox{16cm}{\centering
    {\large Bernhard Egger$^{1,2}$ and Skylar Sutherland$^{1,3}$ and Safa C. Medin$^1$ and Joshua Tenenbaum$^1$}\\
    {\normalsize
    $^1$Massachusetts Institute of Technology\\
    $^2$Friedrich-Alexander-University Erlangen-Nuremberg\\
    $^3$Yale University\\
    bernhard.egger@fau.de
    }}
}

\begin{document}

\ifFGfinal
\thispagestyle{empty}
\pagestyle{empty}
\else
\author{Anonymous FG2021 submission\\ Paper ID \FGPaperID \\}
\pagestyle{plain}
\fi
\maketitle

\begin{abstract}

3D Morphable Models are a class of generative models commonly used to model faces. They are typically applied to ill-posed problems such as 3D reconstruction from 2D data. Several ambiguities in this problem's image formation process have been studied explicitly. We demonstrate that non-orthogonality of the variation in identity and expression can cause \textit{identity-expression ambiguity} in 3D Morphable Models, and that in practice expression and identity are far from orthogonal and can explain each other surprisingly well. Whilst previously reported ambiguities only arise in an inverse rendering setting, identity-expression ambiguity emerges in the 3D shape generation process itself. We demonstrate this effect with 3D shapes directly as well as through an inverse rendering task, and use two popular models built from high quality 3D scans as well as a model built from a large collection of 2D images and videos. We explore this issue's implications for inverse rendering and observe that it cannot be resolved by a purely statistical prior on identity and expression deformations.

\end{abstract}

\section{INTRODUCTION}
3D Morphable Models (3DMMs) of faces were first introduced by Blanz and Vetter \cite{blanz1999morphable} as a statistical model of 3D shape and albedo applicable in an ill-posed inverse rendering setting, and have since been applied in face recognition, entertainment, medicine, forensics, cognitive science, neuroscience, and psychology \cite{egger20203d}. Although 3DMMs originally only modeled neutral faces, they were later extended to incorporate an additional statistical model of facial expressions \cite{amberg2008expression}. In this work we focus on this extension and investigate 3DMMs that aim to separate expression from facial identity.

The seminal work proposing a separate expression model states the following \cite{amberg2008expression}: ``We have observed, that the identity space contains a bit of expressions, mainly smiles, which is due to the difficulty of acquiring perfectly neutral expressions''. It seems the authors were aware that identity and expression are in fact ambiguous in practice, but they view the effect as negligible. In this paper we study identity-expression ambiguity and demonstrate how strong the effect can actually be in state-of-the art 3DMMs. We show not only that the identity space contains expressions to some degree, but also that the expression space can recover many identity features. By studying two of the most popular 3DMMs derived from high-quality 3D-scanned data \cite{FLAMESiggraphAsia2017,BFMgerig2018morphable}, as well as a more recent model learned completely from 2D data \cite{COMPLETEtewari2020learning}, we demonstrate that this effect is not dependent on the specific choice of model, and so is not caused by e.g. registration artifacts.  Rather, it is an intrinsic property of 3DMMs and/or the population-wide distribution of identity and expression.

We limit our study to linear models. Whilst there exist nonlinear models that aim at explicit disentanglement of identity and expression components \cite{abrevaya2019decoupled} they are likely limited in the same way via the training data and the natural distribution of faces which might be at the core of the demonstrated ambiguity.

\subsection{Related Work}
The modeling assumptions underlying 3DMMs have previously been of interest. One core modeling assumption is that 3DMMs separate shape and albedo features. Schumacher et al. demonstrated that this assumption might ease modeling, but does not hold in practice \cite{schumacher2015exploration}. Two works also explicitly studied ambiguities arising in the image formation process of 3DMMs, namely perspective face shape ambiguity \cite{smith2016perspective} and illumination-albedo ambiguity \cite{egger2017semantic}. In contrast, identity-expression ambiguity arises not from 3D-to-2D rendering but rather from the 3D shape generation process itself, and so remains an issue even if the 3D shape is known.

\begin{figure*}[ht]
\includegraphics[width=\textwidth]{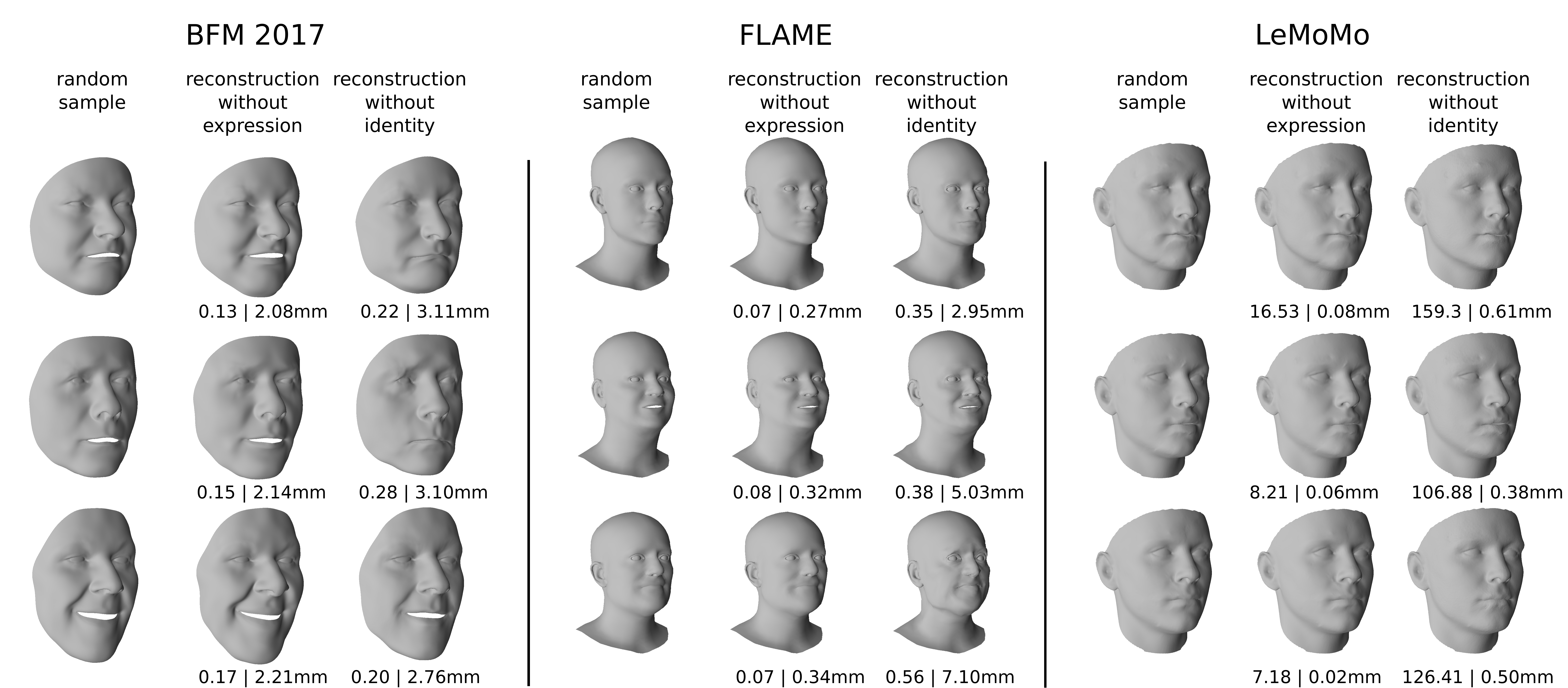}
\caption{Identity-only and expression-only reconstructions of synthetic faces with random identity and expression parameters. We observe that the identity model can reconstruct expression well. Although the expression model seems more constrained, it can still recover shape features that are likely determined by bone structure (i.e. identity) rather than facial expressions. We show the magnitude of the parameters indicating the prior probability of the reconstruction ($\ell_2$ norm divided by amount of parameters; lower is more likely), as well as each reconstruction's shape error in millimeters.}
\label{fig:3Dalphabeta}
\end{figure*}

\begin{figure}
\includegraphics[width=\columnwidth]{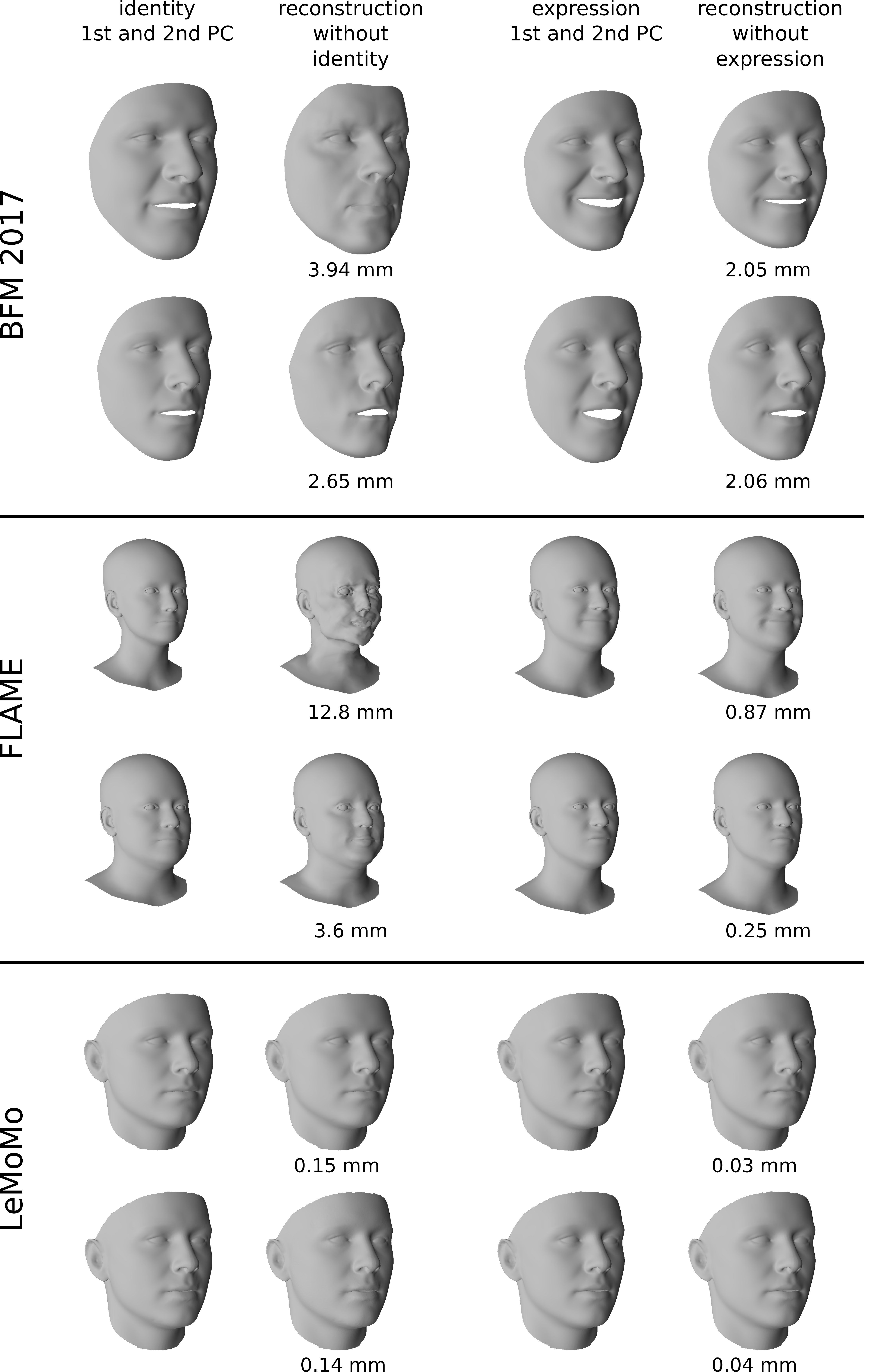}
\caption{Expression-only and identity-only reconstructions of the first 2 identity and expression principal components, respectively. We again show the shape error in millimeters of each reconstruction. Our projection is not regularized (parameters are not constrained to be likely under the prior), this leads to a noisy expression-only reconstruction of the FLAME model's first identity principal component, the overall shape is nevertheless roughly reconstructed.}
\label{fig:3DalphaORbeta}
\end{figure}

\begin{figure}
\includegraphics[width=\columnwidth]{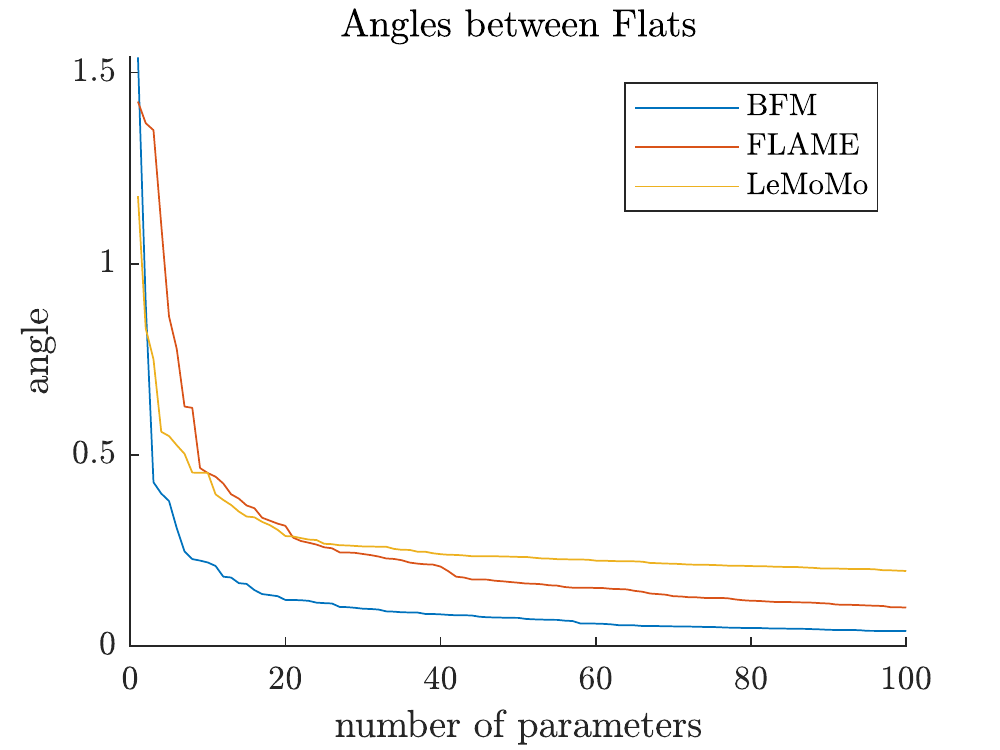}
\caption{The smallest principal angle between the identity and expression shape models' subspaces, as a function of the number of principal components included. An angle of $0$ means that the two subspaces have nontrivial intersection and an angle of $\nicefrac{\pi}{2}$ means the subspaces are orthogonal. We observe that, in all evaluated models, even the first principal components are not orthogonal, and orthogonality drops quickly as more principal components are included.}
\label{fig:anglesbetweenflats}
\end{figure}

\begin{figure*}
\includegraphics[width=\textwidth]{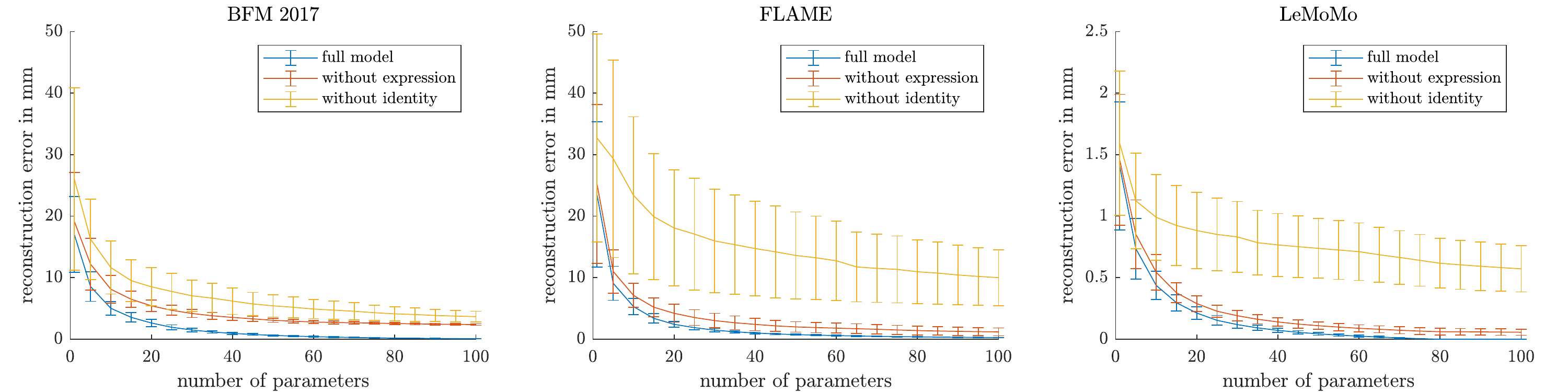}
\caption{Reconstruction quality as a function of the number of parameters used. For the full model, the number of parameters refers to the number of parameters in the identity and expression models individually. We observe that the reconstructions produced using only identity or expression parameters are lower-quality than the reconstructions produced by the full model; nevertheless, the identity-only reconstructions can capture significant expression variation. We further observe that, particularly with the Basel Face Model, the expression model can also explain identity variation. Note that the errors are not comparable between the different models as the covered head/face region differs substantially. Also note the different scale of the LeMoMo model, arising from the model's lower variance.}
\label{fig:nParams}
\end{figure*}

\begin{figure*}
\includegraphics[width=\textwidth]{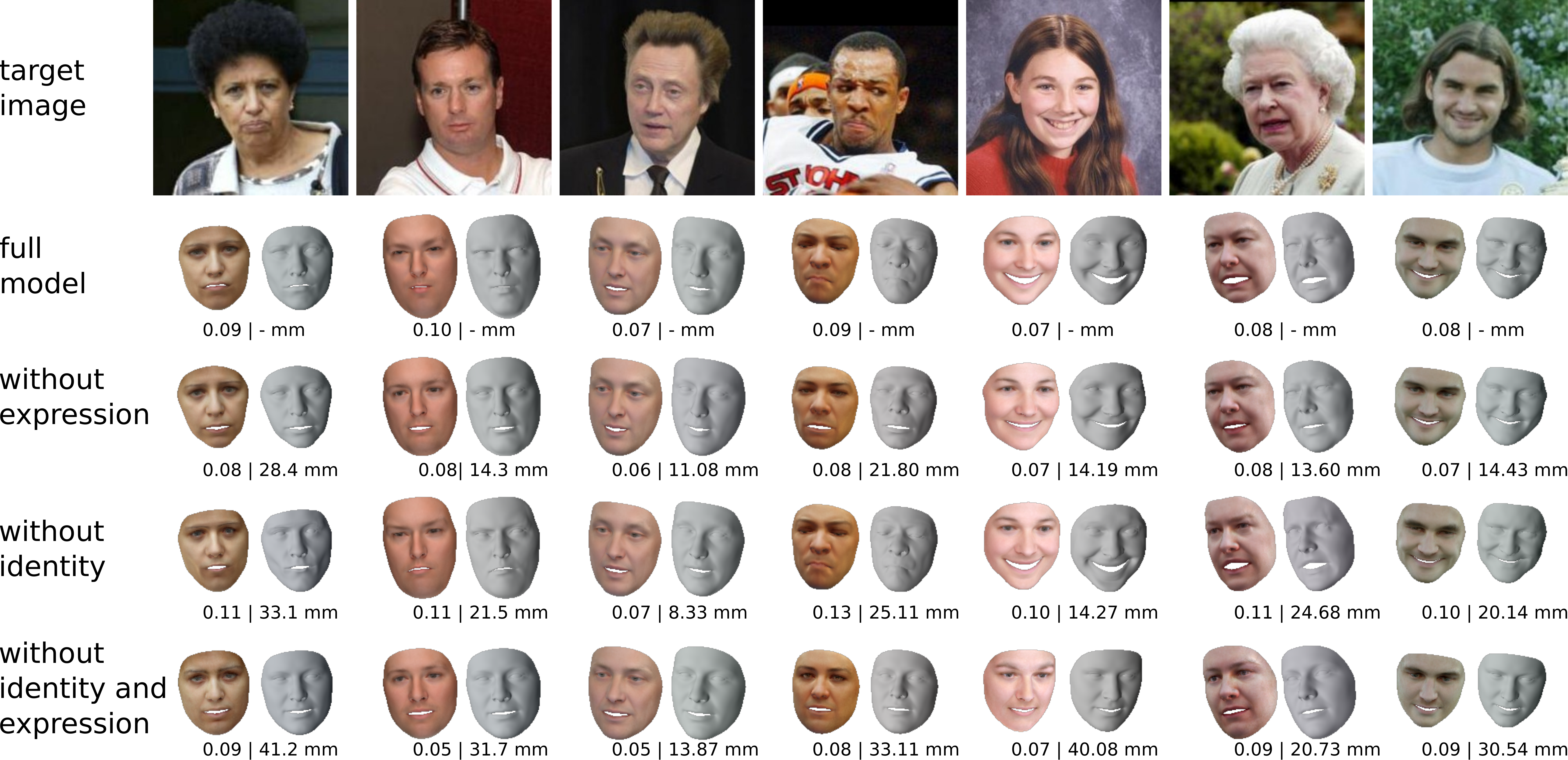}
\caption{Inverse rendering results demonstrating identity-expression ambiguity with real images from the Labeled Faces in the Wild dataset \cite{huang2008labeled} using the 2017 Basel Face Model. We observe that the identity model, expression model, and full model yield very similar image reconstructions and comparable 3D shapes. We also include inference results with neither the identity nor expression models to show the effect of pose, albedo and illumination. We again show the magnitude of the identity, expression and albedo parameters, as well as the difference in reconstructions relative to the full model.}
\label{fig:inverseRendering}
\end{figure*}

\section{METHODS}
In this work we use existing methods to estimate 3DMM parameters for identity and expression from a 3D shape based on \cite{amberg2008expression} or in an inverse rendering setting from a 2D image based on \cite{schonborn2017markov}. Although 3DMMs typically contain identity and expression shape models as well as an albedo model, in this work we focus on the shape parts of the model.

\cite{amberg2008expression} formulates a shape 3DMM with an expression model as a mean shape represented as a vector $\mu \in \mathbb{R}^n$, along with identity and expression matrices $M_{id} \in \mathbb{R}^{n \times m}$ and $M_{exp} \in \mathbb{R}^{n \times k}$, respectively, whose columns represent blendshapes.  An instance $f$ of the 3DMM can be written as $f = \mu + M_{id} \alpha_{id} + M_{exp} \alpha_{exp}$ for some parameters $\alpha_{id}$ and $\alpha_{exp}$.  So long as the columns of $M$ are linearly independent, this equation uniquely determines $\alpha_{id}$ and $\alpha_{exp}$ given $f$. \cite{amberg2008expression} simply assumes linear independence of identity and expression. Since in practice $m + k \ll n$, linear independence of identity and expression is mathematically almost certain.

However, no real face will lie in the measure-$0$ support of a 3DMM. Given realistic noise assumptions, linear near-dependence between identity and expression, or more generally non-orthogonality, introduces ambiguities. Specifically, projection and conversion of 3D face shapes into identity and expression latents is not a measure-preserving process, and so a small amount of noise in the face shape representation can translate into a large amount of noise in latent-space. This is analyzed and proven mathematically in Section~\ref{sec:proof}.

In this paper we seek to investigate the potential impact of this effect in practice. We therefore investigate how much of overall face variation can be explained by identity or expression alone. We study this with both 3D-to-3D and 2D-to-3D reconstruction experiments.  For our 3D-to-3D experiments we use the correspondence between the meshes to obtain the 3DMM latents as described above. For our 2D-to-3D experiments, we use a probabilistic inference technique proposed by \cite{schonborn2017markov}. This technique enables us to simply set the identity or expression parameters to 0 (leading the average of the statistical model) and account for the missing expression or identity deformations using the remaining model parameters.

\section{EXPERIMENTS AND RESULTS}
We conduct two types of experiments. Our first experiments attempt to reconstruct 3D shapes, while our second experiments attempt to infer 3D reconstructions from 2D images. For both experiments we try to explain expression variation with identity parameters and vice-versa identity variation with expression parameters.

We conduct our experiments using the 2017 Basel Face Model (BFM) \cite{BFMgerig2018morphable} as well as the FLAME model \cite{FLAMESiggraphAsia2017}, which are two of the most popular 3DMMs that are publicly available and derived from high quality 3D scans. Those two models were built by separate teams of researchers from datasets with no known overlap. Both models were constructed using scans with controlled facial expressions so as to separate identity from expression as well as possible.

In addition to those two models built from 3D data, we also evaluate a model learned from 2D data. Tewari et al. proposed a 3DMM learned from a 3D template along with 2D images and videos \cite{COMPLETEtewari2020learning}. As this model (LeMoMo) uses video for identity supervision and only a very small set of neutral faces to learn the identity model, it offers valuable insights in how dependent the observed effects are on the model synthesis process.

\subsection{Identity Expression Ambiguity on 3D shapes}
In our first experiment, we sample random identity and expression parameters $\alpha$ and $\beta$ to obtain a random 3D face. The random samples are drawn separately for each model as the following steps of the analysis are calculated in closed form under the correspondence assumption which does not hold between models. We then try to explain the full deformation with identity or expression only. In addition to the qualitative reconstruction results we also measure the average mesh distance to see how accurate our reconstructions are, and include the likelihood of the reconstruction's parameters according to the 3DMMs' statistical models.  The results are visualized in Figure~\ref{fig:3Dalphabeta}. We observe that the reconstructions in both settings can be achieved with reasonably low parameter weights and very close reconstruction results, both perceptually and in terms of reconstruction error in mm. 

In a second experiment we repeat the procedure from the first experiment but generate random 3D faces differently---we vary only the first two principal components of either the identity or expression model and set the other components to the mean. The goal of this experiment is to demonstrate that identity-expression ambiguity can arise without the interplay of expression and identity parameters and is also not an effect solely attributable to later noisy model parameters.  The results are visualized in Figure~\ref{fig:3DalphaORbeta}.

In a third experiment we investigate how orthogonal the subspaces spanned by the identity and expression basis matrices are based on the angles between flats measure \cite{jiang_angles_1996, shonkwiler_poincare_2009}, also known as principal angles. The angle is computed with the method of \cite{shonkwiler_poincare_2009} (see also \cite{mohammadi_principal_2014}). These results are visualized in Figure~\ref{fig:anglesbetweenflats}. We chose the angles between flat measure because it has an intuitive interpretation in terms of how the two spaces relate to each other and offers an interpretable quantitative measure of the subspaces' degree of non-orthogonality. In our case, the spaces are spanned by principal components. We observe that the identity and expression subspaces are highly non-orthogonal.

In a fourth experiment we aim to understand if early and later principal components affect identity-expression ambiguity differently; do later principal components lead to good reconstruction results, or do early and late components contribute equally ambiguity? To study this we compare reconstructions based on both the identity and expression models and compare them to reconstructions based solely on one model. For this experiment we produce reconstructions of the same 100 random meshes per model and per number of parameters and visualize the results in Figure~\ref{fig:nParams}.

\subsection{Identity Expression Ambiguity in Inverse Rendering}
Whilst we have already shown that identity-expression ambiguity occurs in 3D, it is obvious that this will also occur in a setting where the 3DMM is used for a downstream task. We however want to understand the implications for inverse rendering tasks and see how it could affect reconstruction quality, and if the different reconstructions vary in their likelihood. Would we be able to perceive that something is going wrong by looking at the reconstruction results? And can we resolve the ambiguity in an inverse rendering task using the 3DMM's statistical prior? For this experiment we reconstruct images from the Labeled Faces in the Wild dataset \cite{huang2008labeled} and perform inference using only identity or expression to study how well they can explain each other. We also perform inference with both identity \textit{and} expression kept fixed so as to assess the contribution of albedo, illumination and pose. The face reconstruction results can be found in Figure~\ref{fig:inverseRendering}. We observe that the reconstructions based on full shape, identity only or expression only appear very similar. We also observe that we cannot rule out one of the reconstructions solely by enforcing the identity or expression prior since the magnitude of the parameters are very close.

\section{DISCUSSION}
We demonstrate the extent and significance of identity-expression ambiguity in 3DMMs. The effect is surprisingly strong when taking into account that the models were carefully built with strong identity supervision to disentangle identity and expression. Whilst in our experiments we demonstrate the effect in an extreme setting by attempting to explain the full variation of faces with with either identity or expression alone, the effect in real-world contexts, while likely more subtle, could nevertheless introduce severe bias in applications focusing on identity or expression. In practical settings we utilize the prior of the identity and expression models during inference and would produce the reconstruction with the most probable mix of identity and expression parameters. This was shown to be sufficient for expression normalization \cite{egger2017probabilistic} and face recognition under expressions \cite{BFMgerig2018morphable} and demonstrates how important a statistical prior (through implicit or explicit modeling) and probabilistic inference is in 3DMMs. We also show, however, that such a prior does not seem to fully resolve identity-expression ambiguity in inverse rendering as for some images the reconstructions based on solely expression or identity are nearly as likely as the reconstruction based on both identity and expression.

It is unclear how this ambiguity can be resolved, especially since it is related to the fact that there is no real neutral facial expression. Human face perception seems to similarly struggle to separate identity and expression, especially for unfamiliar faces \cite{young2017recognizing}. The adaptation to familiar faces and the resulting shift of subspaces as reported in \cite{she2021geometry} could be an effect of this ambiguity.

Our interpretation of our results is that the intra-person variation at a population level heavily overlaps with inter-person variation---this raises the question of whether facial expressions should be modeled as identity-dependent. While current 3DMMs try to model them independently---which we show may be unfeasible---it may remain possible to incorporate explicit models of the correlations between expression and identity.

\section*{ACKNOWLEDGEMENTS} 

    This work was funded by the DARPA Learning with Less Labels (LwLL) program (Contract No: FA8750-19-C-1001), the DARPA Machine Common Sense (MCS) program (Award ID: 030523-00001) and by the Center for Brains, Minds and Machines (CBMM) (NSF STC award CCF-1231216). B.Egger was supported by a PostDoc Mobility Grant, Swiss National Science Foundation P400P2\_191110.

{\small
\bibliographystyle{ieee}
\bibliography{egbib}
}

\section{PROOF}\label{sec:proof}

Following \cite{amberg2008expression}, we previously formulated a shape 3DMM with an expression model as a mean shape $\mu \in \mathbb{R}^n$, an identity matrix $M_{id} \in \mathbb{R}^{n \times m}$, and an expression matrix $M_{exp} \in \mathbb{R}^{n \times k}$.  We more realistically frame the problem of converting a shape $f \in \mathbb{R}^n$ into a 3DMM's latents $\alpha_{id}(f) \in \mathbb{R}^m$ and $\alpha_{exp}(f) \in \mathbb{R}^k$ as finding the closest face (in a least-squares sense) $\phi(f) \in \mathbb{R}^n$ of the form $\mu + M_{id} \alpha_{id}(f) + M_{exp} \alpha_{exp}(f)$, where $\alpha_{id}(f)$ and $\alpha_{exp}(f)$ are the identity and expression latents inferred from $f$.  Let us define
\[M = \begin{bmatrix} M_{id} & M_{exp} \end{bmatrix} \text{ and } \alpha(f) = \begin{bmatrix} \alpha_{id}(f) \\ \alpha_{exp}(f) \end{bmatrix}.\]
Then we may more simply write $\phi(f)$ as $\mu + M \alpha(f)$.

Now, suppose that our observation of the 3D face shape is somewhat noisy.  In this case, rather than considering only a single face shape $f$, we wish to consider the set $S \subset \mathbb{R}^n$ of all possible face shapes $f$ within $\epsilon$ (in the $\ell_2$ norm) of some reference face $f_0$.  $\alpha(S) \subset \mathbb{R}^{m + k}$ is then the set of all latents we might infer from faces $f \in S$.  We can then ask what the measure of $\alpha(S)$ (which we write as $|\alpha(S)|$) is.  Let us assume without loss of generality that $m \ge k$; if $k < m$, then we may simply swap identity and expression.  We here prove that the principal angles between the 3DMM's identity and expression subspaces are, in increasing order, $\theta_1, \ldots, \theta_k$, then
\[|\alpha(S)| \propto \prod_{i = 1}^k \frac{1}{\sin \theta_i}\]
where the constant of proportionality depends only on $m$, $k$, and $\epsilon$.

\begin{proof}
Let $f \in S$.  We may then write $f$ as $f_0 + \delta$ with $||\delta|| < \epsilon$.  Let $Q \in \mathbb{R}^{n \times (m + k)}$ be a matrix whose columns form an orthonormal basis for the space spanned by the columns of $M_{id}$ and $M_{exp}$.  The support of the 3DMM is then the affine space of points of the form $\mu + Q y$ for $y \in \mathbb{R}^{m + k}$.  The projection of $f$ into this hyperplane is
\begin{align*}
    \phi(f) &= Q Q^{\top} (f - \mu) + \mu\\
            &= Q Q^{\top} (f_0 + \delta - \mu) + \mu\\
            &= \phi(f_0) + Q Q^{\top} \delta.
\end{align*}
$Q Q^{\top} \delta$ is the projection of $\delta$ into the column space of $Q$ (i.e. the set of points of the form $Q y$ for some $y \in \mathbb{R}^{m + k}$).  Given that the columns of $Q$ are orthonormal, we observe that if we define $\beta(f) = Q^{\top} \delta$, then $\beta(S)$ is simply a ball of radius $\epsilon$ around the origin in $\mathbb{R}^{m + k}$.  Let $|\beta(S)| = \mu_0$.  We note that $\mu_0$ only depends on $\epsilon$ and $m + k$.  Now, we observe that
\begin{align*}
    \phi(f) &= \phi(f_0) + Q \beta(f)\\
            &= Q Q^{\top} (f_0 - \mu) + \mu + Q \beta(f)\\
            &= Q (Q^{\top} (f_0 - \mu) + \beta(f)) + \mu.
\end{align*}
So if we let $\gamma(f) = Q^{\top} (f_0 - \mu) + \beta(f)$, then $\mu + Q \gamma(f) = \phi(f) = \mu + M \alpha(f)$.  Furthermore, $\gamma(f)$ is simply a translation of $\beta(f)$, so $\gamma(S)$ must also have measure $\mu_0$.

Clearly $M \alpha(f) = Q \gamma(f)$.  The columns of $Q$ are orthonormal, so $Q^{\top} Q = I_{m + k}$, where $I_j$ is the $j$-by-$j$ identity matrix for any $j$.  Thus $Q^{\top} M \alpha(f) = \gamma(f)$.  Assuming that identity and expression are linearly independent, $Q^{\top} M$ has full rank, so $\alpha(f) = (Q^{\top} M)^{-1} \gamma(f)$.  Since $(Q^{\top} M)^{-1}$ is a linear transformation, it follows that
\[|\alpha(S)| = \mu_0 |\det((Q^{\top} M)^{-1})| = \frac{\mu_0}{|\det(Q^{\top} M)|}.\]

Now, we note that $M_{id}$ and $M_{exp}$ are not arbitrary matrices; they are computed via principal component analysis (PCA) and their columns are orthogonal \cite{amberg2008expression}, and here we rescale them to be orthonormal.  We further note that this derivation works for \textit{any} matrix $Q$ whose columns form an orthonormal basis for the space spanned by the columns of $M$.  Since the columns of $M_{id}$ are orthonormal, we may thus choose the first $m$ columns of $Q$ to be the columns of $M_{id}$, giving us $Q = \begin{bmatrix} M_{id} & Q_{exp} \end{bmatrix}$ for some matrix $Q_{exp}$.  Then
\[Q^{\top} M = \begin{bmatrix} M_{id}^{\top} \\ Q_{exp}^{\top} \end{bmatrix} \begin{bmatrix} M_{id} & M_{exp} \end{bmatrix} = \begin{bmatrix} M_{id}^{\top} M_{id} & M_{id}^{\top} M_{exp} \\ Q_{exp}^{\top} M_{id} & Q_{exp}^{\top} M_{exp} \end{bmatrix}.\]
The columns of $M_{id}$ are orthonormal, so $M_{id}^{\top} M_{id} = I_m$.  The columns $M_{id}$ are also all orthogonal to the columns of $Q_{exp}$, so $Q_{exp}^{\top} M_{id}$ is the zero matrix.  Thus letting $0_{i \times j}$ represent the $i$-by-$j$ zero matrix,
\[Q^{\top} M = \begin{bmatrix} I_m & M_{id}^{\top} M_{exp} \\ 0_{k \times m} & Q_{exp}^{\top} M_{exp} \end{bmatrix}.\]
This is a block-triangular matrix, so
\[\det(Q^{\top} M) = \det(I_m) \det(Q_{exp}^{\top} M_{exp}) = \det(Q_{exp}^{\top} M_{exp}).\]

We observe that this value depends only on the subspaces spanned by the columns of $M_{id}$ and $M_{exp}$.  Clearly $|\det(Q_{exp}^{\top} M_{exp})|$ depends only on $M_{id}$'s column space and not on its specific columns, since $Q_{exp}^{\top}$ is defined purely in terms of $M_{id}$'s column space.  As for $M_{exp}$, we note that for any matrix $A$ whose columns form an orthonormal basis for the column space of $M_{exp}$, there exists some change-of-basis matrix $C$ such that $A C = M_{exp}$.  Since $M_{exp}$ and $A$ have orthonormal columns,
\[C^{\top} C = C^{\top} I_k C = C^{\top} A^{\top} A C = M_{exp}^{\top} M_{exp} = I_k.\]
Thus $C$ is an orthogonal matrix and so
\begin{align*}
    |\det(Q_{exp}^{\top} M_{exp})|  &= |\det(Q_{exp}^{\top} A C)|\\
                                    &= |\det(Q_{exp}^{\top} A)| |\det(C)|\\
                                    &= |\det(Q_{exp}^{\top} A)|.
\end{align*}
So the value of $|\det(Q_{exp}^{\top} M_{exp})|$ depends only on the column spaces of $M_{id}$ and $M_{exp}$, i.e. the identity and expression subspaces.  This means we may arbitrarily choose the columns of $M_{id}$ and $M_{exp}$ so long as they form orthonormal bases for the identity and expression subspaces respectively.

We note that by the definition of principal angles, there exist unit vectors $v$ and $w$ in the identity and expression subspaces, respectively, such that the angle between $v$ and $w$ is $\theta_1$; and moreover, there do \textit{not} exist vectors $v'$ and $w'$ in the identity and expression subspaces such that the angle between $v'$ and $w'$ is less than $\theta_1$.  We also observe that the normalized projection of $w$ into the space orthogonal to the identity subspace, i.e.
\[u = \frac{w - M_{id} M_{id}^{\top} w}{||w - M_{id} M_{id}^{\top} w||},\]
is a unit vector in the column space of $Q_{exp}$.  Geometrically, we may observe that $v$ must be the normalized projection of $w$ into the identity subspace, so
\[v = \frac{M_{id} M_{id}^{\top} w}{||M_{id} M_{id}^{\top} w||}.\]
This means that $u$, $v$, and $w$ are coplanar, and we observe geometrically that $w$ lies between $v$ and $u$; thus the angle between $u$ and $w$ is $\frac{\pi}{2} - \theta_1$.  Since $u$ and $w$ are unit vectors, this means that $u \cdot w = \cos (\frac{\pi}{2} - \theta_1) = \sin \theta_1$.

As previously noted, $|\det(Q_{exp}^{\top} M_{exp})|$ depends only on the identity and expression subspaces, and $Q_{exp}$ was arbitrarily chosen so that the columns of $\begin{bmatrix} M_{id} & Q_{exp}\end{bmatrix}$ form an orthonormal basis for column space of $M$.  Thus we may choose the first column of $M_{id}$ to be $v$, the first column of $M_{exp}$ to be $w$, and the first column of $Q_{exp}$ to be $u$.

To complete the proof, we prove by induction on $k$ that
\[|\det(Q_{exp}^{\top} M_{exp})| = \prod_{i = 1}^k \sin \theta_i.\]

If $k = 1$, then the only column of $Q_{exp}$ is $u$ and the only column of $M_{exp}$ is $w$, and $Q^{\top} M_{exp}$ is a $1$-by-$1$ matrix, so
\[|\det(Q^{\top} M_{exp})| = |u \cdot w| = \sin \theta_1.\]
Now, suppose that $k > 1$.  Let $m_{exp, i}$ be the $i$th column of $M_{exp}$.  Since $M_{exp}$ has orthonormal columns, $m_{exp, i} \cdot w = 0$ for any $i > 1$.  We now argue that $m_{exp, i}$ must also be orthogonal to $v$.

Since $w$ and $m_{exp, i}$ are unit vectors and orthogonal to each other, we may write any unit vector that is coplanar with $m_{exp, i}$ and $w$ as $r(\psi) = \cos \psi w + \sin \psi m_{exp, i}$.  If we let
\[\psi_0 = \arctan\left(-\frac{v \cdot w}{v \cdot m_{exp, i}}\right),\]
and $r_0 = r(\psi_0)$, then
\begin{align*}
    v \cdot r_0 &= \cos \psi_0 (v \cdot w) + \sin \psi_0 (v \cdot m_{exp, i})\\
                &= \cos \psi_0 (v \cdot w + \tan \psi_0 (v \cdot m_{exp, i}))\\
                &= \cos \psi_0 \left(v \cdot w - \frac{v \cdot w}{v \cdot m_{exp, i}} v \cdot m_{exp, i}\right)\\
                &= 0.
\end{align*}
Let $r_1 = r(\psi_0 + \frac{\pi}{2})$.  Then since $w$ and $m_{exp, i}$ are orthogonal unit vectors,
\begin{align*}
    r_0 \cdot r_1   &= \cos \psi_0 \cos\left(\psi_0 + \frac{\pi}{2}\right) + \sin \psi_0 \sin\left(\psi_0 + \frac{\pi}{2}\right)\\
                    &= -\cos \psi_0 \sin \psi_0 + \sin \psi_0 \cos \psi_0\\
                    &= 0.
\end{align*}
Thus $r_0$ and $r_1$ are orthogonal.  They are also clearly coplanar with $w$ and $m_{exp, i}$, so there must exist some $\psi_1$ such that $w = \cos \psi_1 r_0 + \sin \psi_1 r_1$.  But then
\[v \cdot w = \cos \psi_1 v \cdot r_0 + \sin \psi_1 v \cdot r_1 = \sin \psi_1 v \cdot r_1.\]
By assumption there is no vector $w'$ in the expression subspace which forms an angle $\theta < \theta_1$ with $v$.  Clearly $\theta_1 \in (0, \nicefrac{\pi}{2}]$.  Since $\cos$ is a decreasing function on this interval, this means that there cannot be any unit vector $w'$ in the expression subspace with $v \cdot w' = \cos \theta > \cos \theta_1 = v \cdot w$.  However, if $\psi_1 \ne \pm\nicefrac{\pi}{2}$, then $|\sin \psi_1| < 1$, so $|v \cdot r_1| > v \cdot w$; thus either $v \cdot r_1 > v \cdot w$ or $v \cdot (-r_1) > v \cdot w$.  But $r_1$ and $-r_1$ are by construction unit vectors in the column space of $M_{exp}$, i.e. the expression subspace.  This is a contradiction.  Thus $\psi_1 = \pm \nicefrac{\pi}{2}$ and so $w = \pm r_1$.  As previously noted, $m_{exp, i}$ is a unit vector orthogonal to $r_1 = \pm w$; so is $r_0$.  Since $r_0$, $r_1$ and $m_{exp, i}$ are coplanar, this means that $m_{exp, i} = \pm r_0$.  But $v \cdot r_0 = 0$, so $m_{exp, i} \cdot v = 0$ as well.

Let $M_{exp}' \in \mathbb{R}^{n \times (k - 1)}$ be $M_{exp}$ without its first column, i.e. the matrix whose columns are $m_{exp, 2}, \ldots, m_{exp, k}$; similarly, let $Q_{exp}' \in \mathbb{R}^{n \times (k - 1)}$ be $Q_{exp}$ without its first column.  Then
\[Q_{exp}^{\top} M_{exp} = \begin{bmatrix} u \cdot w & u^{\top} M_{exp}' \\ Q_{exp}'^{\top} w & Q_{exp}'^{\top} M_{exp}' \end{bmatrix}.\]

We observe that $u$ may be written as a linear combination of $v$ and $w$.  For any $i > 1$, we have shown that $m_{exp, i} \cdot v = m_{exp, i} \cdot w = 0$, so $m_{exp, i} \cdot u = 0$ as well.  Thus $M_{exp}'^{\top} u$ is the zero matrix.  So
\[Q_{exp}^{\top} M_{exp} = \begin{bmatrix} \sin \theta_1 & 0_{1 \times (k - 1)} \\ Q_{exp}'^{\top} w & Q_{exp}'^{\top} M_{exp}' \end{bmatrix}.\]
This is a block-triangular matrix, so
\[|\det(Q_{exp}^{\top} M_{exp})| = \sin \theta_1 |\det(Q_{exp}'^{\top} M_{exp}')|.\]
Now, we observe that the columns of $M_{exp}'$ form an orthonormal basis for the space of all vectors in the expression space orthogonal to $w$; similarly, the columns of $Q_{exp}'$ form an orthonormal basis for the space of all vectors in the column-space of $Q_{exp}$ orthogonal to $u$.  Let $M_{id}' \in \mathbb{R}^{n \times (m - 1)}$ be defined as $M_{id}$ without its first column; the columns of $M_{id}'$ then form an orthonormal basis for the space of all vectors in the identity space orthogonal to $v$.  As we have shown, all the columns of $M_{exp}'$ are orthogonal to $v$, and a similar argument likewise shows that all the columns of $M_{id}'$ are orthogonal to $w$.  So the column space of $M' = \begin{bmatrix} M_{id}' & M_{exp}' \end{bmatrix}$ is the space of all vectors in the column space of $M$ that are orthogonal to both $v$ and $w$, and by extension to $u$ as well (since $u$ is a linear combination of $v$ and $w$).  Let $Q' = \begin{bmatrix} M_{id}' & Q_{exp}' \end{bmatrix}$.  Then the columns of $Q'$ must be orthonormal (since they are simply a subset of the columns of $Q$), and similarly they a form a basis for the space of all vectors in the column space of $M$ that are orthogonal to both $v$ and $u$, and by extension $w$.  Furthermore, by the definition of principal angles \cite{jiang_angles_1996}, the principal angles between the column spaces of $M_{id}'$ and $M_{exp}'$ are $\theta_2, \ldots, \theta_k$.  So by the inductive hypothesis,
\[|\det(Q_{exp}'^{\top} M_{exp}')| = \prod_{i = 2}^k \sin \theta_k.\]
Thus
\[|\det(Q_{exp}^{\top} M_{exp})| = \prod_{i = 1}^k \sin \theta_k\]
and so
\[|\alpha(S)| = \mu_0 \prod_{i = 1}^k \frac{1}{\sin \theta_k}.\]
\end{proof}

One practical implication of this is that since $0 \le \sin \theta_i \le 1$ for any $i$,
\[|\alpha(S)| \ge \frac{\mu_0}{\sin \theta_1}.\]
Thus the smallest principal angle offers a lower bound on the amount of numerical instability in the computation of $\alpha(f)$.
\end{document}